\documentclass[10pt,twocolumn,letterpaper]{article}

\usepackage{cvpr}
\usepackage{times}
\usepackage{epsfig}
\usepackage{graphicx}
\usepackage{amsmath}
\usepackage{amssymb}
\usepackage{algorithm2e}
\usepackage{comment}
\usepackage{arydshln}

\usepackage[breaklinks=true,bookmarks=false]{hyperref}

\DeclareMathOperator{\GRU}{GRU}

\cvprfinalcopy % *** Uncomment this line for the final submission

\def\cvprPaperID{1517} % *** Enter the CVPR Paper ID here

\begin{document}

%%%%%%%%% TITLE
\title{Human Motion Anticipation with Symbolic Label}

\author{Julian Tanke\\
University of Bonn\\
{\tt\small tanke@iai.uni-bonn.de}
% For a paper whose authors are all at the same institution,
% omit the following lines up until the closing ``}''.
% Additional authors and addresses can be added with ``\and'',
% just like the second author.
% To save space, use either the email address or home page, not both
\and
Andreas Weber\\
University of Bonn\\
{\tt\small weber@cs.uni-bonn.de}
\and
J\"urgen Gall\\
University of Bonn\\
{\tt\small gall@iai.uni-bonn.de}
}

\maketitle
%\thispagestyle{empty}

%%%%%%%%% ABSTRACT
\begin{abstract}
   Anticipating human motion depends on two factors: the past motion and the person's intention.
While the first factor has been extensively utilized to forecast short sequences of human motion, the second one remains elusive.
In this work we approximate a person's intention via a symbolic representation, for example fine-grained action labels such as walking or sitting down.
Forecasting a symbolic representation is much easier than forecasting the full body pose with its complex inter-dependencies.
However, knowing the future actions makes forecasting human motion easier.
We exploit this connection by first anticipating symbolic labels and then generate human motion, conditioned on the human motion input sequence as well as on the forecast labels.
This allows the model to anticipate motion changes many steps ahead and adapt the poses accordingly.
We achieve state-of-the-art results on short-term as well as on long-term human motion forecasting.
\end{abstract}

%%%%%%%%% BODY TEXT
\section{Introduction}

Anticipating human motion is highly relevant for many interactive activities such as sports, manufacturing, or navigation~\cite{paden2016survey} and significant progress has been made in forecasting human motion~\cite{fragkiadaki2015recurrent,gopalakrishnan2019neural,gui2018adversarial,jain2016structural,li2018convolutional,martinez2017human,pavllo2019modeling}.
Conceptually, human motion anticipation depends on two factors: the past motion and the intention of the person.
The first factor naturally lends itself to being modelled recursively, which has been successfully exploited by many recent works~\cite{gui2018adversarial,li2018convolutional,liu2019towards,martinez2017human}.
These models achieve impressive results in forecasting short time horizons less than a half second but exhibit common issues such as converging to mean poses and noise accumulation for longer time horizons.
While one emerging solution to this problem is the use of adversarial training strategies~\cite{gui2018teaching,li2018convolutional,ruiz2018human} to regularize the model output, we propose to model the intent of the person as a complementary approach to solve this problem. 
The motivation behind this is that for longer time horizons it is easier to forecast the intention or actions rather than the full body pose with all its intricacies. However, if one knows the future actions, forecasting the human motion becomes an easier task.

\begin{figure}[t]
  \begin{center}
  \includegraphics[width=0.8\linewidth]{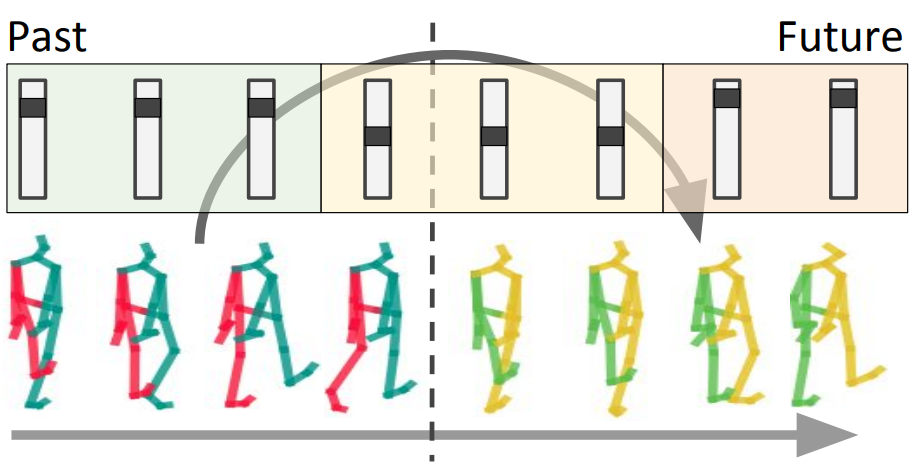}
\end{center}
\caption{
Given a human motion input sequence, represented as blue-red skeletons, our model forecasts the future human motion, represented as yellow-green skeletons. However, instead of predicting the future poses directly from the past poses (bottom), we additionally anticipate a symbolic representation for the motion (top). The human poses are forecast based on the past human poses and the forecast symbolic labels.
}
\label{fig:idea}
\end{figure}

\begin{figure*}[t]
  \begin{center}
  \includegraphics[width=0.9\linewidth]{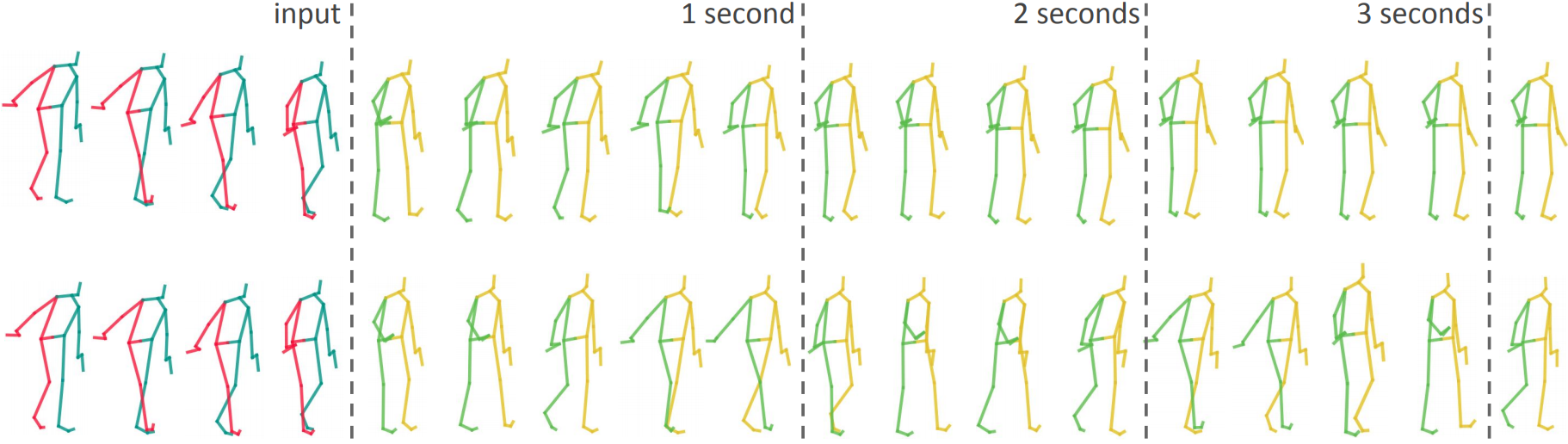}
\end{center}
\caption{
Qualitative results of our model: the blue-red skeletons represent the model input while the yellow-green skeletons are the model output.
The top row represents a sequence which was generated without symbolic labels. After $1$ second the motion freezes into a mean pose.
The bottom row represents a sequence generated with symbolic labels where sensible motion is retained for several seconds.
}
\label{fig:qual}
\end{figure*}

\begin{figure}[t]
  \begin{center}
  \includegraphics[width=1\linewidth]{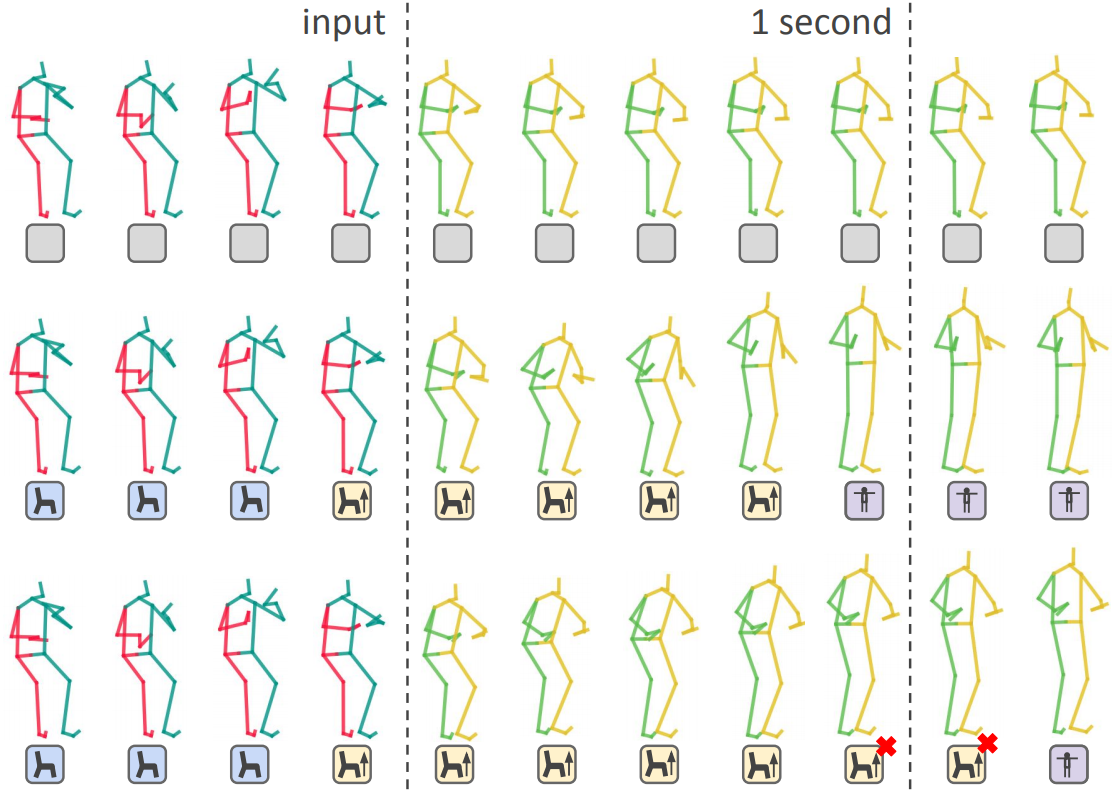}
\end{center}
\caption{
Qualitative results of our model with action anticipation:
the middle row represents the ground truth data with the associated human-labeled action labels below. The sequence captures a person in the process of standing up.
The top row sequence was generated without action labels: instead of initiating the stand-up motion it remains sitting on a chair.
The bottom row was generated with action label forecasting: the model correctly detects the initiation of standing up and forecasts action labels and poses.
Errors in the label forecasting are marked with a red \texttt{x}.
}
\label{fig:sitting}
\end{figure}

In order to approximate the intention of a person, we use a symbolic representation. This symbolic representation can be considered as fine-grained actions like walking or standing that on one hand abstracts the human motion by a categorical representation and on the other hand is fine-grained enough such that each symbol represents a motion that is not too diverse. Given such symbolic representation, our model forecasts the human motion not directly from the past human poses, but it first forecasts the high-level symbolic representation from which it then generates the future human poses, conditioned on the past motion, as it is illustrated in Figure \ref{fig:idea}. This means that the forecast human poses depend not only on the past motion sequence but also on the past, current and future symbolic labels. 
This allows the model to anticipate motion changes many steps ahead resulting in smooth transitions.
Furthermore, it allows anticipation of longer time horizons as the categorical labels help the model to generate realistic poses instead of degenerating to a mean pose.
Qualitative results of anticipating long time horizons can be seen in Figure \ref{fig:qual} while Figure \ref{fig:sitting} provides an example of generating smooth transitions by forecasting future labels. 

Our model is trained end-to-end and consists of several components. It first extracts from the observed human poses the symbolic labels. These are then forecast by a sequence-to-sequence module. Finally, the poses are forecast by a recurrent network that takes the past observed human poses and the past and future symbolic labels as input. This means that the network already knows when forecasting the first pose what actions will happen next.  

We evaluate our approach for human motion forecasting on the Human 3.6M~\cite{h36m_pami} dataset. On this dataset, our approach achieves state-of-the-art results for time horizons up to four seconds. We also thoroughly evaluate the impact of the symbolic representation.

% ---------- M O D E L  F I G U R E ---------------------- (START)
\begin{figure*}[t]
  \begin{center}
  \includegraphics[width=0.7\linewidth]{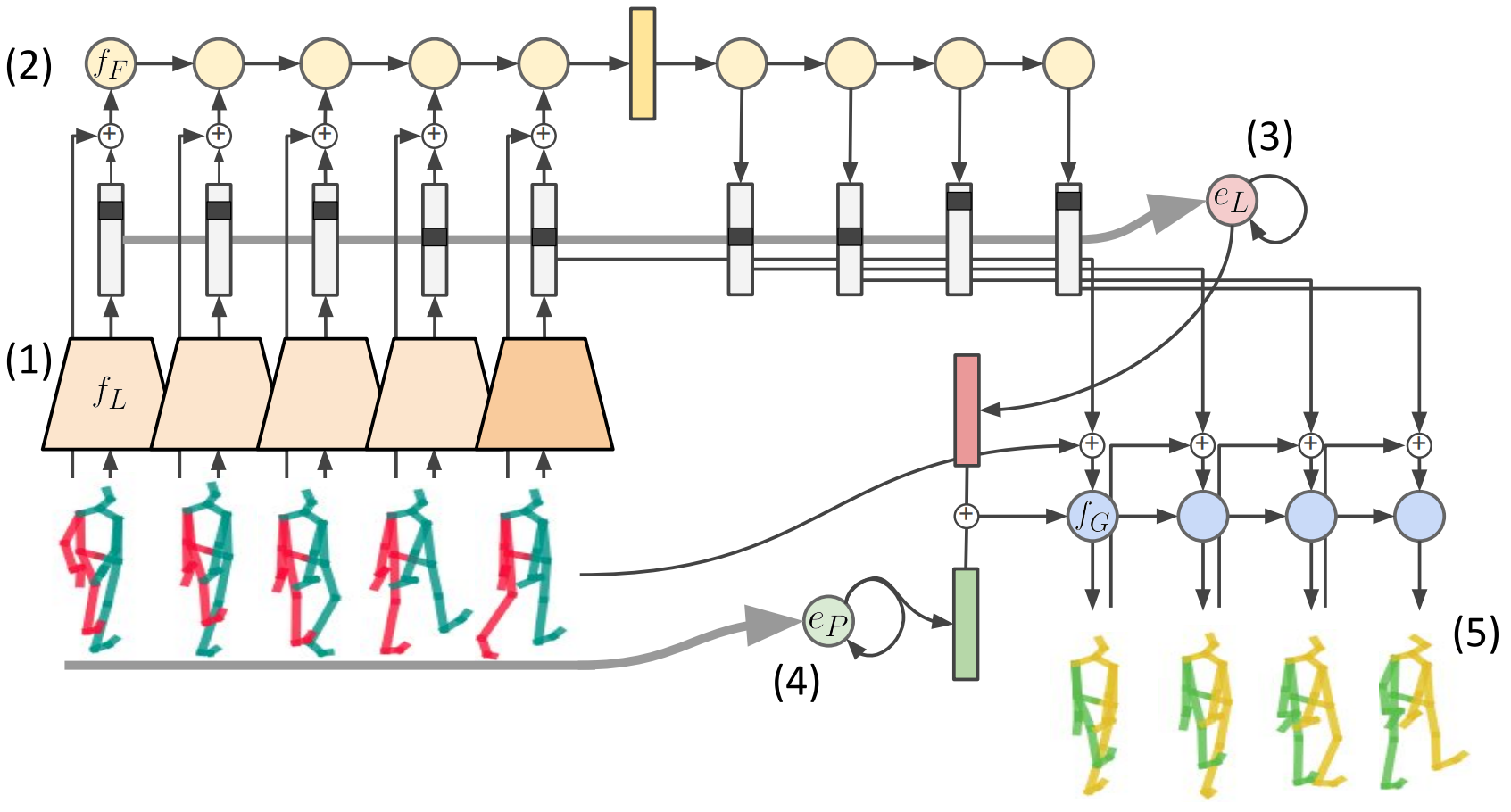}
\end{center}
\caption{
Network overview: (1) The symbolic label extraction module $f_L$ takes the observed human poses $\mathbf{x}_1^t$ as input and predicts frame-wise probabilities for the symbolic labels $\hat{\mathbf{c}}_1^t$. (2) For each frame, the probabilities $\hat{c}_{\tau}$ are concatenated with the human pose vector $x_{\tau}$. A sequence-to-sequence module then forecasts the symbolic labels $\hat{\mathbf{c}}_{t+1}^T$. It consists of a recurrent encoder $e_F$, which maps the sequence to a vector, followed by a recurrent decoder $d_F$, which predicts $\hat{\mathbf{c}}_{t+1}^T$. (3) Past and future symbolic labels $\hat{\mathbf{c}}_{1}^T$ are mapped to a single vector $v_L$ by the recurrent label encoder $e_L$. (4) Similarly, the past poses $\hat{\mathbf{x}}_{1}^t$ are mapped to a single vector $v_P$ by the recurrent pose encoder $e_P$. (5) The hidden states $v_L$ and $v_P$ are concatenated and passed to the recurrent pose generator $f_G$. It recurrently estimates the poses $\hat{\mathbf{x}}_{t+1}^T$ from the previous estimated pose and the current symbolic label.   
}
\label{fig:model}
\end{figure*}

\begin{figure}[t]
  \begin{center}
  \includegraphics[width=0.99\linewidth]{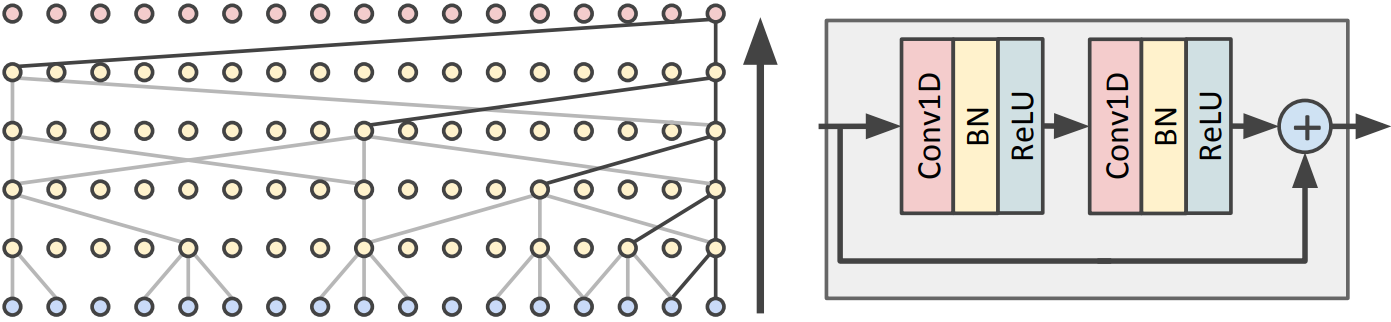}
\end{center}
\caption{
Graphical representation of the symbolic label extraction module $f_L$ and its residual block.
}
\label{fig:tcncell}
\end{figure}
% ---------- M O D E L  F I G U R E ---------------------- (END)

\section{Related Work}

In recent years deep neural networks have been used to synthesize and anticipate 3D human poses from motion capture data.
Significant progress has been made in synthesizing human motion from data~\cite{Hodgins:2017:ipend,Hodgins:2017:DOE,Hodgins:2017:shedulefrag,starke2019nsm}.
However, these models require control input and do not forecast human motion.
Holden et al.~\cite{holden2016deep} show that autoencoders can be utilized to learn a human motion manifold.
B\"utepage et al.~\cite{butepage2017deep} extend this idea by embedding the skeletal hierarchy structure of the data into the model.
Similarly, structural RNNs~\cite{jain2016structural} encode the hierarchy utilizing RNNs.
Encoder-Recurrent-Decoder (ERD)~\cite{fragkiadaki2015recurrent} auto-regressively forecasts human motion by utilizing an encoder-decoder structure for modeling human poses and an LSTM for temporal modelling.
ERD suffers from error accumulation and requires careful training by adding Gaussian noise to the input data.
Auto-conditioned recurrent networks~\cite{zhou2018auto} extend the learning procedure by not relying on teacher forcing but by feeding back the network output to the next time step prediction in fixed interval steps, making the model less susceptible to error accumulation.
Residual sequence-to-sequence architectures~\cite{martinez2017human} model first-order motion derivatives using a sequence-to-sequence model~\cite{sutskever2014sequence} popularized in machine translation, however, with shared weights between encoder and decoder.
QuaterNet~\cite{pavllo:quaternet:2018} replaces the exponential map representation of previous works by a quaternion representation, which does not suffer from common 3D rotational problems such as gimbal locks.
The loss is calculated in 3D Euclidean space by using forward kinematics and joints are weighted according to their importance in the kinematic chain.
A similar approach is utilized in Hierarchical Motion Recurrent networks~\cite{liu2019towards} where a novel RNN structure is proposed which better represents skeletal structures.
Graph-convolutional neural networks~\cite{mao2019learning} can be utilized to learn human motion in trajectory space rather than in pose space, which enables the network to model long-range dependencies beyond that of the kinematic tree.

Recently, models based on adversarial training gained some attention:
Convolutional Sequence-to-Sequence models~\cite{li2018convolutional} utilize a convolutional encoder-decoder structure with adversarial loss to prevent overfitting.
Adversarial Geometry-Aware encoder-decoder (AGED)~\cite{gui2018adversarial} utilize two adversarial losses: one to tackle motion discontinuity, which is a common problem in previous models, and one to ensure that realistic motion is generated.
On top of that, the geodesic rather than the Euclidean distance is used as reconstruction loss.
Spatio-Temporal Motion Inpainting (STMI-GAN)~\cite{ruiz2018human} frames human motion anticipation as inpainting problem which can be solved using a GAN.
The model learns the joint distribution of body poses and motion, enabling it to hypothesize large portions of missing data.

Recently, anticipating future actions from data~\cite{koppula2015anticipating,lan2014hierarchical} has seen growing interest:
Abu Farha et al.~\cite{abu2018will,farha2019ms} utilize an RNN to forecast action labels from videos.
Ke et al.~\cite{ke2019time} and Gammulle et al.~\cite{gammulle2019forecasting} build on this by utilizing temporal convolutions with time-variable and memory networks, respectively.
These models are capable of forecasting action labels several minutes into the future.

\section{Model}

In this work we address the task of forecasting human motion. This means that we observe 3d human skeletons for $t$ frames, which are denoted by $\mathbf{x}_1^t = (x_1, \hdots, x_{t} ) \in \mathbb{R}^{t\times d}$ and where $d$ is the feature dimension that represents the human pose, and our goal is to forecast the future pose sequence $\mathbf{x}_{t+1}^T$. While previous approaches use a network to anticipate the future poses directly from the past poses, we propose to use a symbolic representation, which abstracts human poses on a higher level, and forecast the symbolic representation as it is illustrated in Figure~\ref{fig:idea}. The forecast poses depend then on the forecast symbolic representation of the motion and the observed human poses in the past. In our experimental evaluation, we will show that the symbolic representation substantially improves the quality of the forecast human motion for time horizons beyond one second in the future.  

Figure~\ref{fig:model} illustrates the network, which consists of three components. The pose label predictor $f_L$, which is described in Section~\ref{sec:label}, infers from the observed poses $\mathbf{x}_1^t$ the symbolic labels $\hat{\mathbf{c}}_1^t$. The network component $f_F$, which is described in Section~\ref{sec:labelfor}, uses the observed poses and inferred symbolic labels to forecast the symbolic labels for the future frames $\hat{\mathbf{c}}_{t+1}^T$. The last component $f_G$, which is described in Section~\ref{sec:posefor}, forecasts the human poses $\hat{\mathbf{x}}_{t+1}^{T}$ based on the inferred and forecast symbolic labels $\hat{\mathbf{c}}_{1}^T$ and the observed poses $\mathbf{x}_1^t$.

% ============================================================
% P O S E  L A B E L  P R E D I C T O R
% ============================================================
\subsection{Pose Label Predictor}\label{sec:label}

For the pose label predictor $f_L$, we use a temporal convolutional network which takes as input a human motion sequence $\mathbf{x}_1^t$ and infers the respective symbolic labels $\mathbf{c}_1^t$, \ie,
\begin{equation}
\mathbf{c}_1^t = f_L(\mathbf{x}_1^t).
\end{equation}
The model has a kernel size of $3$, with exponentially increasing dilation factors, similar to WaveNet~\cite{oord2016wavenet}.
We utilize $5$ layers which results in a receptive field of $65$ frames.
Each layer is composed of the standard TCN residual block~\cite{bai2018empirical} without Dropout, which is shown in Figure \ref{fig:tcncell}.
A $1\times 1$ convolution is added to the first layer to match the number of hidden units in the residual block.
As intermediate loss for $f_L$, we use the categorical cross-entropy:
\begin{equation}
\mathcal{L}_{f_L} = \frac{1}{t} \sum_{\tau=1}^t \sum_{j=1}^{n} c_{\tau j} \log (\hat{c}_{\tau j}) 
\end{equation}
where $n$ is the total number of discrete symbolic labels, $c_{\tau j}$ denotes the ground-truth value of the $j$-th class at time step $\tau$, and $\hat{c}_{\tau j}$ denotes the predicted probability of the $j$-th class at time step $\tau$.

% ============================================================
%  L A B E L  F O R E C A S T I N G
% ============================================================
\subsection{Label Forecasting}\label{sec:labelfor}

For label forecasting, we use a sequence-to-sequence model, which consists of a recurrent encoder $e_F$ and a recurrent decoder $d_F$, as shown in Figure~\ref{fig:model}. The recurrent encoder consists of GRU units that take the previous estimated hidden state $v_{\tau-1}$ and the current pose $x_{\tau}$ concatenated with the probabilities of the symbolic labels $\hat{c}_{\tau}$ as input:        
\begin{equation}
v_{\tau} = \GRU(v_{\tau-1}, x_{\tau} \oplus \hat{c}_{\tau}).
\end{equation}
For $\tau=t$, the entire sequence information is then encoded by the vector 
\begin{equation}
v_{t} = e_F(\mathbf{x}_1^t,\hat{\mathbf{c}}_1^t).
\end{equation}
The decoder $d_F$ also consists of GRU units that start with $v_t$ and forecast for $t < \tau \leq T$ the symbolic labels, \ie,   
\begin{equation}
(v_{\tau},c_{\tau}) = \GRU(v_{\tau-1})
\end{equation}
where an additional softmax layer is applied to $c_{\tau}$ to obtain the probabilities for the symbolic labels $\hat{c}_{\tau}$. The label forecasting component $f_F$ is thus the combination of the recurrent encoder and decoder:
\begin{equation}
\hat{\mathbf{c}}_{t+1}^T = f_F(\mathbf{x}_1^t,\hat{\mathbf{c}}_1^t) = d_F(e_F(\mathbf{x}_1^t,\hat{\mathbf{c}}_1^t)).
\end{equation}
Similar to $f_L$, the intermediate loss is defined as follows:
\begin{equation}
    \mathcal{L}_{f_F} = \frac{1}{T-t}
    \sum_{\tau=t+1}^T \sum_{j=1}^n
    c_{\tau j} \log ( \hat{c}_{\tau j} )
\end{equation}
where $\hat{c}_{\tau j}$ is the forecast probability of label $j$ at time step $\tau$.

% ============================================================
%  P O S E  F O R E C A S T I N G  G E N E R A T O R
% ============================================================
\subsection{Forecasting Human Motion}\label{sec:posefor}

For human motion forecasting, we also use a recurrent network $f_G$, as shown in Figure~\ref{fig:model}.
The hidden state of the network $p_{\tau}$ is initialized by two recurrent encoders $e_L$ and $e_P$, which use GRU units as all encoders in the network. 
The encoder $e_L$ takes as input the inferred and forecast symbolic labels $\hat{\mathbf{c}}_1^T$ and encodes the entire sequence information in a single vector.
The motivation behind this is that $f_G$ then already knows the sequence of abstract motion it needs to generate, allowing for smooth transitions between different motions. 
The second encoder $e_P$ takes as input the human motion input sequence $\mathbf{x}_1^t$ and encodes the observed human pose sequence in a single vector. The vectors of both encoders are then concatenated to obtain
\begin{equation}
    p =  e_L( \hat{\mathbf{c}}_1^T ) \oplus e_P(\mathbf{x}_1^t).  
\label{eq:eq8}
\end{equation}

The recurrent network $f_G$ consists of GRU units and recursively forecasts the human poses for the frames ${t < \tau \leq T}$. It takes for each frame $\tau$ the previous estimated pose $\hat{x}_{\tau-1}$ concatenated with the forecast symbolic label $\hat{c}_{\tau}$ as input:     
\begin{equation}
    (p_\tau, \hat{x}_{\tau}) = \GRU(p_{\tau-1}, \hat{x}_{\tau-1} \oplus \hat{c}_{\tau}).
\label{eq:eq9}
\end{equation}
For $\tau = t+1$, $p_{\tau-1}=p$ and $\hat{x}_{\tau-1} = x_{t}$.  
We train our network with the L$2$ loss
\begin{equation}
    \mathcal{L}_{f_G} = \frac{1}{J \cdot (T-t)}
    \sum_{\tau=t+1}^T \sum_{j=1}^J \vert\vert x_{\tau j} - \hat{x}_{\tau j} \vert\vert_2
\end{equation}
where $J$ is the number of joints in the pose and $x_{\tau j}$ and $\hat{x}_{\tau j}$ denote the ground truth and model prediction of joint $j$ at time frame $\tau$, respectively. To obtain a smooth transition between the observed and forecast poses, we use a \textit{warm-up} phase for $f_G$.
This means that the network does not directly start to forecast $\hat{x}_{t+1}$, but it also estimates the latest $w$ observed poses $\hat{\mathbf{x}}_{t-w+1}^t$.

% ============================================================
%  T R A I N I N G  
% ============================================================
\subsection{Implementation Details}

We train our model using the Adam optimizer with a learning rate of $0.0005$ and a batch size of $16$.
After the optimizer converges, we switch to SGD with a learning rate of $0.0001$~\cite{keskar2017improving}. 
We also found that using hard-labels (one-hot encoding) works best for pose forecasting.
During training, we thus feed the ground-truth labels to $f_G$ as well as to $f_F$.
During evaluation, argmax is applied to the softmax outputs and the model operates end-to-end.
For all our experiments, we set the number of hidden units per convolution in $f_L$ to $256$,
the number of hidden units in GRUs $e_F$, $d_F$ and $e_P$ to $512$, the numbers of hidden units in GRU $e_L$ to $256$, and the numbers of hidden units in GRU $f_G$ to $768$ to match the concatenation of $e_L(\mathbf{\hat{c}}_1^T)$ and $e_P(\textbf{x}_1^t)$.
If not otherwise stated, our model uses a warm-up period of $w=24$.
The effect of $w$ is evaluated in Section \ref{sec:warmup}.

% -------- insert here
\begin{table*}[t]
\footnotesize
\begin{center}
\begin{tabular}{c | c c c c c c | c c c c c c }
 &  \multicolumn{6}{c}{Walking} & \multicolumn{6}{c}{Eating} \\
milliseconds & 80 & 160 & 320 & 400 & 560 & 1000 & 80 & 160 & 320 & 400 & 560 & 1000 \\ \hline
ResSup~\cite{martinez2017human}&21.7&38.1&58.9&68.8&79.4&91.6&15.1&28.6&54.8&67.4&82.6&110.8\\
CNN~\cite{li2018convolutional}&21.8&37.5&55.9&63.0&69.2&81.5&13.3&24.5&48.6&60.0&71.8&91.4\\
CNN (3D)~\cite{li2018convolutional}&17.1&31.2&53.8&61.5&59.2&71.3&13.7&25.9&52.5&63.3&66.5&85.4\\
TD~\cite{mao2019learning}&\textbf{8.9}&\textbf{15.7}&\textbf{29.2}&\textbf{33.4}&\textbf{42.3}&\textbf{51.3}&8.8&18.9&39.4&47.2&56.5&\underline{68.6}\\
\hline 
Ours (k=4)&\underline{10.1}&21.0&37.4&40.3&\underline{42.7}&59.6&\underline{8.2}&\underline{17.1}&\underline{34.2}&\underline{42.1}&\underline{54.2}&76.9\\
Ours&10.4&\underline{19.8}&\underline{31.6}&\underline{35.1}&43.7&\underline{54.1}&\textbf{7.6}&\textbf{15.0}&\textbf{30.1}&\textbf{38.2}&\textbf{43.9}&\textbf{60.2}\\
\multicolumn{13}{c}{} \\
 &  \multicolumn{6}{c}{Smoking} & \multicolumn{6}{c}{Discussion} \\
milliseconds & 80 & 160 & 320 & 400 & 560 & 1000 & 80 & 160 & 320 & 400 & 560 & 1000 \\ \hline
ResSup~\cite{martinez2017human}&20.8&39.0&66.1&76.1&89.5&122.6&26.2&51.2&85.8&94.6&121.9&154.3\\
CNN~\cite{li2018convolutional}&15.4&25.5&39.3&44.5&50.3&85.2&23.6&43.6&68.4&74.9&101.0&143.0\\
CNN (3D)~\cite{li2018convolutional}&11.1&21.0&33.4&38.3&42.0&67.9&18.9&39.3&67.7&75.7&84.1&116.9\\
TD~\cite{mao2019learning}&\textbf{7.8}&\underline{14.9}&\underline{25.3}&\underline{28.7}&\textbf{32.3}&\underline{60.5}&\textbf{9.8}&\underline{22.1}&\textbf{39.6}&\textbf{44.1}&\textbf{70.5}&\underline{103.5}\\
\hline 
Ours (k=4)&8.3&15.9&30.0&38.2&49.5&62.8&15.8&32.8&69.7&86.7&106.6&128.2\\
Ours&\underline{8.2}&\textbf{14.6}&\textbf{24.2}&\textbf{27.6}&\underline{33.2}&\textbf{56.8}&\underline{10.6}&\textbf{21.7}&\underline{54.1}&\underline{65.5}&\underline{83.3}&\textbf{93.2}\\
% -------- insert (end)
\end{tabular}
\end{center}
\caption{
3D positional error in \textit{millimeters} for $8$ sub-sequences per action, as defined in the evaluation protocol~\cite{jain2016structural}.
The numbers are taken from~\cite{mao2019learning}.
}
\label{tab:compare_with_trajectory}
\end{table*}

% -------- insert here
\begin{table*}[t]
\footnotesize
\begin{center}
\begin{tabular}{c | c c c c c c | c c c c c c }
 &  \multicolumn{6}{c}{Walking} & \multicolumn{6}{c}{Eating} \\
milliseconds & 80 & 160 & 320 & 400 & 560 & 1000 & 80 & 160 & 320 & 400 & 560 & 1000 \\ \hline
Zero Velocity~\cite{martinez2017human}&27.2&50.8&86.7&100.1&117.3&120.0&9.7&18.4&33.2&40.1&52.1&\underline{73.8}\\
ResSup~\cite{martinez2017human}&24.6&44.9&73.3&82.6&92.3&105.2&18.6&35.1&60.5&70.3&85.8&112.7\\
Lie~\cite{liu2019towards}&28.9&43.5&74.3&82.7&91.7&127.2&16.7&29.6&51.4&59.4&73.0&112.0\\
CNN~\cite{li2018convolutional}&23.1&42.2&68.9&77.1&86.8&95.6&13.7&25.8&44.7&52.6&66.3&94.2\\
QuaterNet~\cite{pavllo2019modeling}&13.7&27.5&50.7&59.0&71.0&89.5&9.3&19.6&39.4&48.6&66.3&101.2\\
TD~\cite{mao2019learning}&12.8&23.8&\underline{38.7}&\underline{43.6}&\underline{50.1}&\underline{58.2}&\underline{8.1}&\underline{16.9}&\underline{32.7}&\underline{39.8}&\underline{52.1}&74.9\\
\hline 
Ours (k=4)&\underline{12.0}&\underline{22.6}&39.0&44.9&52.2&59.4&9.3&17.7&33.7&40.9&52.5&76.3\\
Ours&\textbf{11.4}&\textbf{20.3}&\textbf{31.8}&\textbf{35.9}&\textbf{41.6}&\textbf{51.8}&\textbf{7.9}&\textbf{15.5}&\textbf{28.9}&\textbf{35.3}&\textbf{46.6}&\textbf{67.3}\\
\multicolumn{13}{c}{} \\
 &  \multicolumn{6}{c}{Smoking} & \multicolumn{6}{c}{Discussion} \\
milliseconds & 80 & 160 & 320 & 400 & 560 & 1000 & 80 & 160 & 320 & 400 & 560 & 1000 \\ \hline
Zero Velocity~\cite{martinez2017human}&21.0&40.0&69.5&80.7&96.1&115.9&27.1&51.3&88.7&103.1&124.4&154.0\\
ResSup~\cite{martinez2017human}&22.0&41.9&73.2&85.1&102.9&132.6&30.2&56.8&97.2&112.7&139.5&179.0\\
Lie~\cite{liu2019towards}&16.6&30.5&55.0&64.8&80.9&116.4&25.8&47.4&80.3&92.4&112.8&146.3\\
CNN~\cite{li2018convolutional}&16.1&29.7&51.0&59.5&72.6&99.3&23.8&43.5&73.7&86.0&107.2&139.0\\
QuaterNet~\cite{pavllo2019modeling}&9.5&19.6&38.8&47.3&61.9&91.2&\underline{14.9}&\underline{31.0}&\underline{63.8}&78.5&102.5&138.4\\
TD~\cite{mao2019learning}&\textbf{8.6}&\textbf{17.5}&\underline{32.9}&\underline{39.5}&\underline{50.4}&\underline{71.6}&\textbf{12.9}&\textbf{27.5}&\textbf{56.1}&\textbf{68.6}&\textbf{88.7}&\textbf{117.8}\\
\hline 
Ours (k=4)&\underline{9.4}&\underline{17.9}&34.1&41.5&54.0&76.0&17.6&35.0&66.8&79.7&100.3&131.0\\
Ours&10.1&18.6&\textbf{31.8}&\textbf{37.0}&\textbf{45.2}&\textbf{64.9}&18.3&35.7&66.0&\underline{78.3}&\underline{96.3}&\underline{123.6}\\
% -------- insert (end)
\end{tabular}
\end{center}
\caption{
3D positional error in \textit{millimeters} for $256$ sub-sequences per action to better represent the test distribution, as defined in the evaluation protocol~\cite{pavllo2019modeling}.
}
\label{tab:shortterm}
\end{table*}

% -------- insert here
\begin{table*}[t]
\tiny
\begin{center}
\begin{tabular}{c | c c c c c c | c c c c c c | c c c c c c |}
 &  \multicolumn{6}{c}{Directions} & \multicolumn{6}{c}{Greeting} & \multicolumn{6}{c}{Phoning} \\
milliseconds & 80 & 160 & 320 & 400 & 560 & 1000 & 80 & 160 & 320 & 400 & 560 & 1000 & 80 & 160 & 320 & 400 & 560 & 1000 \\ \hline
Zero Velocity~\cite{martinez2017human}&15.2&28.9&50.7&\underline{59.6}&\underline{74.5}&\underline{105.8}&23.5&44.0&77.1&91.0&113.0&149.0&24.5&46.4&82.5&97.0&119.8&140.4\\
ResSup~\cite{martinez2017human}&24.6&47.1&82.8&96.8&117.6&154.0&36.0&67.1&112.6&129.5&156.7&200.2&24.0&45.1&77.4&89.3&108.6&142.4\\
Lie~\cite{liu2019towards}&19.6&37.8&69.0&81.8&99.8&133.3&30.6&57.5&99.5&114.7&137.5&173.1&19.5&37.1&65.9&75.9&92.2&131.7\\
CNN~\cite{li2018convolutional}&19.4&36.8&64.7&76.7&94.5&128.7&29.8&55.3&91.7&105.4&126.9&161.0&18.1&33.5&59.0&69.5&88.5&126.5\\
QuaterNet~\cite{pavllo2019modeling}&11.8&26.5&56.0&68.8&87.0&122.5&20.0&41.9&79.4&94.0&118.2&158.2&\underline{12.0}&24.2&47.6&58.4&78.3&120.2\\
TD~\cite{mao2019learning}&\textbf{9.8}&\textbf{22.4}&\underline{48.9}&60.0&77.3&107.5&\underline{17.9}&\underline{36.7}&\underline{70.4}&\underline{83.2}&\underline{103.7}&\underline{134.7}&\textbf{10.1}&\textbf{20.8}&\textbf{41.2}&\textbf{50.6}&\underline{68.0}&104.6\\
\hline 
Ours (k=4)&14.0&28.2&56.7&69.3&89.2&119.6&21.5&43.1&83.2&98.2&118.8&147.4&12.2&\underline{23.4}&44.9&54.5&69.8&\underline{102.7}\\
Ours&\underline{11.6}&\underline{22.5}&\textbf{41.8}&\textbf{50.5}&\textbf{66.2}&\textbf{97.4}&\textbf{16.0}&\textbf{30.4}&\textbf{56.3}&\textbf{68.2}&\textbf{87.0}&\textbf{124.3}&12.8&24.0&\underline{43.7}&\underline{51.9}&\textbf{65.4}&\textbf{93.2}\\
\multicolumn{19}{c}{} \\
 &  \multicolumn{6}{c}{Posing} & \multicolumn{6}{c}{Purchases} & \multicolumn{6}{c}{Sitting} \\
milliseconds & 80 & 160 & 320 & 400 & 560 & 1000 & 80 & 160 & 320 & 400 & 560 & 1000 & 80 & 160 & 320 & 400 & 560 & 1000 \\ \hline
Zero Velocity~\cite{martinez2017human}&25.4&48.8&88.4&103.9&129.6&188.3&30.5&56.2&92.0&104.9&126.2&172.2&18.5&35.5&63.6&75.6&97.0&142.4\\
ResSup~\cite{martinez2017human}&33.7&65.3&117.1&137.2&169.3&224.7&33.0&60.6&101.6&116.5&139.8&175.8&28.7&53.7&91.8&106.0&128.8&174.6\\
Lie~\cite{liu2019towards}&26.1&50.2&93.1&110.9&141.5&204.6&27.6&52.2&88.8&102.9&125.6&164.0&18.5&35.0&62.6&74.5&95.7&138.3\\
CNN~\cite{li2018convolutional}&24.1&46.5&85.9&103.0&133.3&193.6&27.0&49.0&81.5&94.6&117.2&156.4&18.5&34.2&60.5&72.0&92.6&133.3\\
QuaterNet~\cite{pavllo2019modeling}&\underline{15.5}&\underline{34.2}&74.3&92.8&126.1&193.4&\underline{20.3}&40.1&76.1&91.6&116.1&159.9&13.2&27.0&55.6&69.2&93.0&140.3\\
TD~\cite{mao2019learning}&\textbf{13.1}&\textbf{30.0}&\textbf{66.0}&\textbf{82.4}&\textbf{112.9}&175.6&\textbf{16.5}&\textbf{35.1}&\textbf{66.9}&\textbf{80.1}&\textbf{99.7}&\textbf{132.1}&\textbf{10.7}&\textbf{22.6}&\textbf{45.6}&\underline{57.0}&\underline{77.5}&\underline{120.6}\\
\hline 
Ours (k=4)&17.2&34.6&\underline{69.5}&\underline{85.6}&\underline{113.1}&\underline{172.4}&20.6&\underline{39.8}&\underline{74.8}&90.6&115.8&\underline{151.9}&\underline{13.2}&\underline{25.4}&49.5&61.1&81.7&126.5\\
Ours&19.2&38.0&73.8&89.3&116.0&\textbf{169.6}&23.8&44.7&76.6&\underline{84.1}&\underline{110.3}&159.1&14.2&27.2&\underline{49.2}&\textbf{56.3}&\textbf{73.9}&\textbf{113.1}\\
\multicolumn{19}{c}{} \\
 &  \multicolumn{6}{c}{Sitting Down} & \multicolumn{6}{c}{Taking Photo} & \multicolumn{6}{c}{Waiting} \\
milliseconds & 80 & 160 & 320 & 400 & 560 & 1000 & 80 & 160 & 320 & 400 & 560 & 1000 & 80 & 160 & 320 & 400 & 560 & 1000 \\ \hline
Zero Velocity~\cite{martinez2017human}&25.5&47.8&83.4&100.3&134.0&187.8&17.9&34.2&63.5&76.9&99.4&150.6&19.1&36.3&64.1&74.8&91.0&119.1\\
ResSup~\cite{martinez2017human}&41.3&76.1&127.2&146.7&180.4&239.3&28.2&53.2&94.0&110.6&137.5&188.5&27.4&51.8&91.2&106.4&129.0&168.9\\
Lie~\cite{liu2019towards}&26.4&50.3&86.8&102.6&131.1&182.7&20.8&37.9&68.6&82.2&106.9&155.3&22.8&42.9&76.7&89.8&110.3&149.5\\
CNN~\cite{li2018convolutional}&26.3&48.2&79.8&93.4&119.2&166.6&18.6&33.9&59.5&71.1&93.0&140.6&20.9&39.1&70.0&82.4&102.1&138.7\\
QuaterNet~\cite{pavllo2019modeling}&\underline{18.9}&38.5&74.1&89.6&117.4&173.7&\underline{11.4}&\underline{23.9}&\underline{50.7}&\underline{63.0}&\underline{84.4}&\underline{128.3}&\underline{12.4}&26.6&56.3&69.8&92.9&130.2\\
TD~\cite{mao2019learning}&\textbf{17.0}&\textbf{34.3}&\textbf{63.4}&\textbf{76.5}&\textbf{101.9}&\textbf{148.0}&\textbf{10.1}&\textbf{21.3}&\textbf{43.6}&\textbf{54.1}&\textbf{74.9}&\textbf{117.8}&\textbf{11.2}&\textbf{23.7}&\underline{48.9}&\underline{60.4}&\underline{79.2}&\underline{108.5}\\
\hline 
Ours (k=4)&25.9&49.3&89.8&111.8&146.9&212.8&12.9&25.8&52.2&65.8&90.7&143.8&12.9&25.4&50.6&62.0&82.3&113.4\\
Ours&19.3&\underline{37.5}&\underline{65.6}&\underline{78.4}&\underline{104.5}&\underline{150.1}&14.3&27.7&54.5&67.5&90.5&144.2&13.2&\underline{25.2}&\textbf{46.1}&\textbf{54.6}&\textbf{67.8}&\textbf{101.0}\\
\multicolumn{19}{c}{} \\
 &  \multicolumn{6}{c}{Walking Dog} & \multicolumn{6}{c}{Walking Together} & \multicolumn{6}{c}{Average} \\
milliseconds & 80 & 160 & 320 & 400 & 560 & 1000 & 80 & 160 & 320 & 400 & 560 & 1000 & 80 & 160 & 320 & 400 & 560 & 1000 \\ \hline
Zero Velocity~\cite{martinez2017human}&52.9&98.2&153.5&165.3&170.5&201.7&21.9&41.8&73.5&85.0&100.2&103.0&24.0&45.2&78.0&90.6&109.7&141.6\\
ResSup~\cite{martinez2017human}&41.2&74.8&119.1&134.4&157.0&195.1&22.4&41.7&68.8&78.6&93.2&113.9&29.1&54.3&92.5&106.8&129.2&167.1\\
Lie~\cite{liu2019towards}&38.2&66.7&106.1&121.4&146.1&180.6&24.6&41.8&68.6&78.3&92.9&129.9&24.2&44.0&76.5&89.0&109.2&149.7\\
CNN~\cite{li2018convolutional}&36.2&63.7&98.7&111.1&130.6&171.7&20.3&37.9&63.7&72.7&84.9&101.7&22.4&41.3&70.2&81.8&101.0&136.5\\
QuaterNet~\cite{pavllo2019modeling}&\underline{25.5}&\underline{49.0}&\underline{85.1}&\underline{98.5}&\underline{123.7}&176.0&11.3&23.8&45.0&53.7&67.7&89.8&\underline{14.6}&30.2&59.5&72.2&93.8&134.2\\
TD~\cite{mao2019learning}&\textbf{24.2}&\textbf{47.4}&\textbf{80.9}&\textbf{93.5}&\textbf{109.0}&\textbf{147.8}&11.3&22.3&40.2&47.0&56.6&66.5&\textbf{13.0}&\textbf{26.8}&\textbf{51.8}&\underline{62.4}&\underline{80.1}&\underline{112.4}\\
\hline 
Ours (k=4)&28.6&52.9&92.8&110.1&134.9&\underline{169.0}&\underline{10.4}&\underline{19.9}&\underline{35.9}&\underline{41.4}&\underline{49.3}&\underline{61.7}&15.8&30.7&58.2&70.5&90.1&124.3\\
Ours&37.0&66.7&101.4&112.4&132.9&172.3&\textbf{9.6}&\textbf{17.8}&\textbf{29.9}&\textbf{34.1}&\textbf{39.2}&\textbf{48.8}&15.9&\underline{30.1}&\underline{53.2}&\textbf{62.3}&\textbf{78.9}&\textbf{112.0}\\
% -------- insert (end)
\end{tabular}
\end{center}
\caption{
3D positional error in \textit{millimeters} for $256$ sub-sequences per action to better represent the test distribution, as defined in the evaluation protocol~\cite{pavllo2019modeling}.
}
\label{tab:shorttermall}
\end{table*}

\begin{table}[t]
\footnotesize
\begin{center}
\begin{tabular}{| c|c|c |c |c | } 
\hline
\multicolumn{1}{| c | }{Model} & \multicolumn{4}{c|}{Short-Term ($0-1s$)} \\
 \hline
 & Walking & Eating & Smoking & Discussion\\
 \hline
 ResSup~\cite{martinez2017human}              & 0.161          & 0.214            & 0.265          & 0.703 \\
 vGRU-r1 (SA)~\cite{gopalakrishnan2019neural} & 0.120          & 0.091            & \textbf{0.052} & 0.258 \\
  GRU-d~\cite{gopalakrishnan2019neural}       & 0.127          & \underline{0.095}& 0.126          & \underline{0.185} \\
 vGRU-d~\cite{gopalakrishnan2019neural}     & \underline{0.117}& 0.121            & 0.084          & 0.194 \\
 \hline
 Ours (exp)                                   & \textbf{0.107} & \textbf{0.076}  & \underline{0.053}& \textbf{0.120} \\
 \hline
 \multicolumn{1}{|c | }{} & \multicolumn{4}{c|}{Medium-Term ($1-2s$)} \\
  \hline
 ResSup~\cite{martinez2017human}              & 0.237          & 0.160          & 0.405          & 0.477 \\
 vGRU-r1 (SA)~\cite{gopalakrishnan2019neural} & 0.194          & 0.093          & 0.079          & 0.375 \\
 GRU-d~\cite{gopalakrishnan2019neural}      & \textbf{0.170} & \underline{0.096}& 0.083          & \underline{0.258} \\
 vGRU-d~\cite{gopalakrishnan2019neural}     & \underline{0.179}& \textbf{0.080} & \underline{0.067}& 0.331 \\
 \hline
Ours (exp)                                     & 0.185          & 0.110          & \textbf{0.033} & \textbf{0.240} \\
 \hline
 \multicolumn{1}{|c | }{} & \multicolumn{4}{c|}{Long-Term ($2-4s$)} \\
  \hline
 ResSup~\cite{martinez2017human}              & 0.549          & 0.754          & 1.403          & 1.245 \\
 vGRU-r1 (SA)~\cite{gopalakrishnan2019neural} & 0.544          & 0.764          & 0.948          & 2.720 \\
 GRU-d~\cite{gopalakrishnan2019neural}   & \underline{0.406}   & 0.332          & 0.723          & \textbf{0.785} \\
 vGRU-d~\cite{gopalakrishnan2019neural}       & \textbf{0.359} & \textbf{0.288} & \textbf{0.577} & 1.001 \\
 \hline
Ours (exp)                                    & 0.488   & \underline{0.305} & \underline{0.708} & \underline{0.858} \\
 \hline
\end{tabular}
\end{center}
\caption{
NPSS~\cite{gopalakrishnan2019neural} at $3$ different time horizons i.e short-term: (0-1 seconds), medium-term (1-2 seconds) and long-term (2-4 seconds).
NPSS is reported for $8$ test sequences per action in Euler space as in \cite{gopalakrishnan2019neural}.
}
\label{tab:npss_scales}
\end{table}

\begin{table}[t]
\scriptsize
\begin{center}
% ========= BEGIN GENERATE (NPSS MAIN TABLE) ==============
\begin{tabular}{|c|c|c|c|c|c| }
\hline
\multicolumn{1}{| c | }{Model} & \multicolumn{5}{c|}{Short-Term ($0-1s$)} \\
 \hline
 & Walking & Eating & Smoking & Discussion & Average \\
\hline
ResSup~\cite{martinez2017human}&0.164&0.058&0.080&0.145&0.143\\
CNN~\cite{li2018convolutional}&0.159&0.050&0.054&\underline{0.113}&0.105\\
QuaterNet~\cite{pavllo2019modeling}&0.120&0.050&0.047&0.117&0.109\\
TD~\cite{mao2019learning}&0.086&\underline{0.036}&0.042&\textbf{0.096}&\underline{0.096}\\
Lie~\cite{liu2019towards}&0.167&0.055&0.060&\underline{0.113}&0.119\\
\hdashline
Ours&\textbf{0.045}&\textbf{0.035}&0.042&0.120&\textbf{0.092}\\
Ours (k=4)&0.068&0.037&\underline{0.038}&0.124&0.108\\
Ours (k=8)&0.063&0.037&\textbf{0.038}&0.117&0.102\\
Ours (k=32)&\underline{0.060}&0.040&0.039&0.123&0.103\\
Ours (k=64)&0.065&0.039&0.041&0.128&0.125\\
Ours (gt)&\textit{0.045}&\textit{0.034}&\textit{0.041}&\textit{0.114}&\textit{0.102}\\
\hline
\multicolumn{1}{| c | }{ } & \multicolumn{5}{c|}{Medium-Term ($1-2s$)} \\
\hline
ResSup~\cite{martinez2017human}&0.218&0.062&0.103&0.143&0.154\\
CNN~\cite{li2018convolutional}&0.220&0.077&0.075&\textbf{0.135}&\textbf{0.130}\\
QuaterNet~\cite{pavllo2019modeling}&0.185&0.080&0.082&0.148&0.162\\
TD~\cite{mao2019learning}&0.356&0.155&0.205&0.248&0.298\\
Lie~\cite{liu2019towards}&0.253&0.087&0.085&0.151&0.170\\
\hdashline
Ours&\textbf{0.064}&\textbf{0.040}&0.077&0.147&\underline{0.134}\\
Ours (k=4)&0.103&0.051&\textbf{0.058}&0.138&0.136\\
Ours (k=8)&0.110&0.054&0.064&0.140&0.135\\
Ours (k=32)&\underline{0.097}&\underline{0.049}&0.062&\underline{0.136}&0.138\\
Ours (k=64)&0.108&0.051&\underline{0.062}&0.139&0.147\\
Ours (gt)&\textit{0.064}&\textit{0.040}&\textit{0.073}&\textit{0.141}&\textit{0.133}\\
\hline
\multicolumn{1}{| c | }{ } & \multicolumn{5}{c|}{Long-Term ($2-4s$)} \\
\hline
ResSup~\cite{martinez2017human}&0.714&0.281&0.312&\textbf{0.507}&0.512\\
CNN~\cite{li2018convolutional}&0.695&0.310&\underline{0.301}&\underline{0.509}&\underline{0.505}\\
QuaterNet~\cite{pavllo2019modeling}&0.804&0.422&0.369&0.565&0.638\\
TD~\cite{mao2019learning}&0.700&0.655&0.871&1.074&1.293\\
Lie~\cite{liu2019towards}&0.915&0.513&0.685&0.721&0.864\\
\hdashline
Ours&\textbf{0.193}&\textbf{0.185}&\textbf{0.296}&0.520&\textbf{0.497}\\
Ours (k=4)&0.371&0.250&0.301&0.518&0.532\\
Ours (k=8)&0.406&0.246&0.325&0.536&0.522\\
Ours (k=32)&\underline{0.336}&\underline{0.233}&0.335&0.529&0.517\\
Ours (k=64)&0.380&0.252&0.327&0.533&0.535\\
Ours (gt)&\textit{0.193}&\textit{0.180}&\textit{0.249}&\textit{0.514}&\textit{0.478}\\
\hline
\end{tabular}
% ========= END GENERATE (NPSS MAIN TABLE) ==============
\end{center}
\caption{
NPSS~\cite{gopalakrishnan2019neural} at $3$ different time horizons i.e short-term: (0-1 seconds), medium-term (1-2 seconds) and long-term (2-4 seconds).
NPSS is reported for $256$ test sequences per action in Euclidean space.
}
\label{tab:npss_main}
\end{table}

% ==========================================================================
% E X P E R I M E N T S
% ==========================================================================
\section{Experiments}

\begin{figure}[t]
  \begin{center}
  \includegraphics[width=1\linewidth]{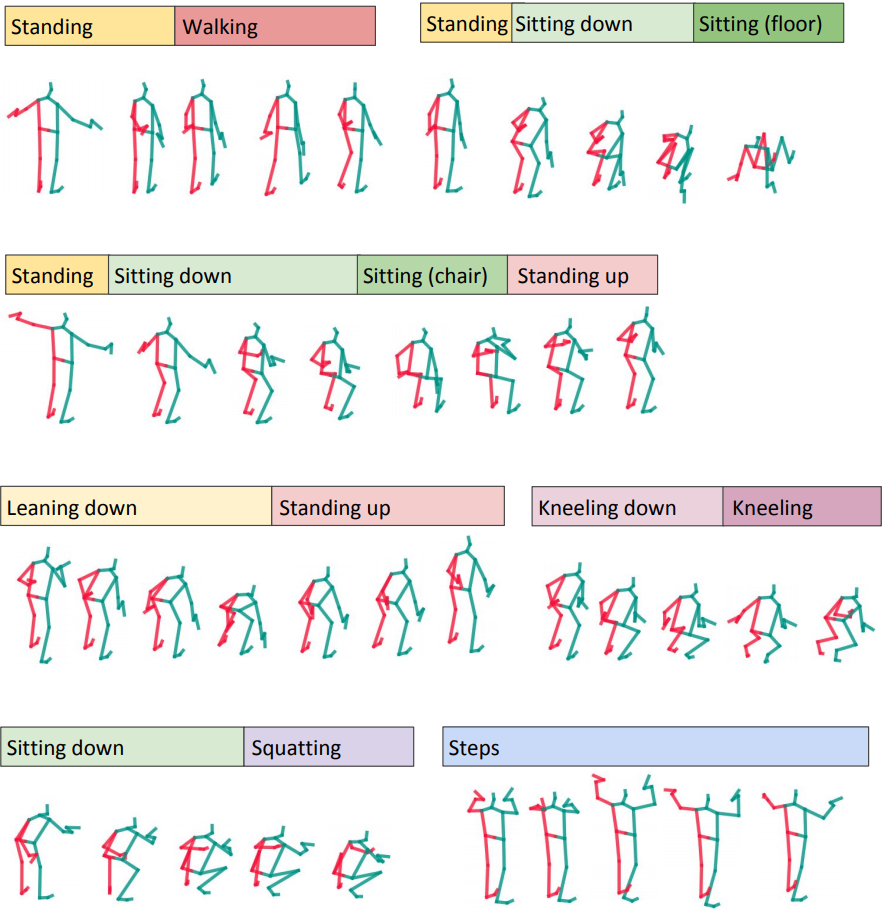}
\end{center}
\caption{
Samples  for  all $11$ action labels obtained by human labeling.
When a person is \textit{Standing}, the legs are firmly placed on the ground.
\textit{Walking} is a directed forward-motion while \textit{Steps} are undirected backwards or side-steps.
Persons may also \textit{Squat}, \textit{Sit on the floor}, \textit{Sit on a chair} or \textit{Kneel}.
We also provide transitional labels.
We distinguish between \textit{Sitting down} and \textit{Kneeling down} as they are visually distinct in the mocap sequences.
}
\label{fig:humanlabels}
\end{figure}

% ================================================
% D A T A
% ================================================
\subsection{Data Representation}
\label{sec:datarep}

For training and evaluation we utilize Human3.6M (H3.6M)~\cite{h36m_pami}, a large-scale human mocap dataset with seven actors performing $15$ different actions such as \textit{walking} and \textit{smoking}.
The dataset is sampled at $50$Hz and has in total $527.599$ frames.
The dataset consists of $210$ sequences which are between $20$ and $127$ seconds long with an average length of $50$ seconds.
As it was proposed in \cite{mao2019learning}, we utilize 3D joint positions rather than an angular representation.
To make our model comparable with models that where trained on rotational representations~\cite{li2018convolutional,liu2019towards,martinez2017human,pavllo2019modeling}, we map an Euler angle representation to 3D joint coordinates using actor S1, as it was done in \cite{mao2019learning}.
Global rotation and translation are removed for this step.

% ============================================================
% F R A M E W I S E  C L A S S  L A B E L S
% ============================================================
\subsection{Frame-wise Symbolic Labels}

For obtaining frame-wise symbolic labels, we apply two approaches:
First, we annotate all $527.599$ frames with one of $11$ action labels: 
\textit{kneeling, kneeling down, leaning down, sitting on chair, sitting down, sitting on floor, squatting, standing, standing up, steps, walking}.
The labels were chosen as they are easy to identify in motion capture data and as occur in all actor sequences.
Figure \ref{fig:humanlabels} describes all human-annotated action labels.
Second, we generate symbolic labels:
For a single frame, we concatenate the pose with the next $10$ poses and apply PCA to reduce the dimension of the feature vector to $32$.
We then use k-means to cluster $k$ symbolic labels.
To obtain labels for the test set, we extract PCA features in the same way and assign them to the label of the closest center.
% More statistics with regards to our labeling process can be found in the supplementary material.

% ============================================================
% S O T A  M O D E L S
% ============================================================
\subsection{Comparison to State-of-the-art}

We compare our model to several recent state-of-the-art methods for which we use their available open-sourced implementations.
For QuaterNet~\cite{pavllo2019modeling,pavllo:quaternet:2018} we use the weights provided by the authors while we retrained the other methods based on the best-performing setups.
The method proposed by Mao et al.~\cite{mao2019learning} (TD) is set apart from the other methods by not being recurrent but instead working with fixed sized input and output sequences.
We thus trained the model to predict $1$ second into the future and recursively apply its full output sequence back as input to forecast longer time horizons.
For works that utilize rotational data~\cite{li2018convolutional,liu2019towards,martinez2017human,pavllo:quaternet:2018}, we apply forward kinematics with actor S1 to obtain the 3D positions of the joints.
The Zero Velocity baseline as well as the trajectory dependency (TD) model~\cite{mao2019learning} utilize 3D pose data directly.

% ================================================
% S H O R T  T E R M
% ================================================
\subsection{Short-term Motion Prediction}
\label{sec:shortterm}

To evaluate short-term motion prediction, we follow the same protocol as described in \cite{mao2019learning} where the L2 distance between ground-truth and predicted 3D joint coordinates are calculated.
For each of the 15 actions of the Human 3.6M dataset, $8$ random sub-sequences are used.
Our results can be seen in Table \ref{tab:compare_with_trajectory} where we achieve highly competitive results.
The highly competitive trajectory dependency (TD) model~\cite{mao2019learning} achieves better results in Walking while we perform best on the Eating sequences.
For the Smoking and Discussion sequences, our model ties with \cite{mao2019learning} but performs better for longer forecast horizons, \ie $1$ second.

Pavllo et al.~\cite{pavllo2019modeling} note that evaluating on $8$ random sub-sequences per action produces high variance when choosing different sequences.
To better represent the test distribution, we also adapt their evaluation protocol where $256$ random sub-sequences per action are chosen.
Under this protocol we achieve state-of-the-art results as can be seen in Tables \ref{tab:shortterm} and \ref{tab:shorttermall}.
Our model is very strong on sequences with repeating patterns such as Walking, Eating or Smoking but also achieves state-of-the-art results on some sequences with high variation such as Greeting, Waiting or Sitting.
As expected, our model outperforms all other methods for time horizons longer than $400$ milliseconds, when averaged over all sequences.

% ================================================
% L O N G  T E R M
% ================================================
\subsection{Long-term Motion Prediction}

Finding good evaluation metrics for long-term human motion forecasting is an open research problem due to its stochasticity.
Recently, phase-based metrics~\cite{gopalakrishnan2019neural,ruiz2018human} have been introduced to evaluate longer time horizons.
They represent the predicted sequences as samples from a probability distribution by casting the them as a pseudo probability using Fourier transform.
However, the data format in these works~\cite{gopalakrishnan2019neural,ruiz2018human} is in Euler angles which we argue is sub-optimal for this approach as the Fourier transform requires its base to be continuous while Euler angles are discontinuous.
For this reason we utilize the 3D joint coordinates, as described in Section \ref{sec:datarep}, in our experiments.

To evaluate long-term motion prediction, we utilize the normalized power spectrum similarity (NPSS)~\cite{gopalakrishnan2019neural}.
It compares a ground-truth sequence with its forecast prediction in the power spectrum, obtained by the Fourier transform.
Under NPSS, periodic motions such as walking will spike at certain frequencies while aperiodic actions such as discussion will show a more uniform spread in the frequency domain.
NPSS normalizes the frequencies so that they sum to $1$, resulting in a form of pseudo probability.
To compare the two distributions, the Wasserstein distance~\cite{rubner2000earth} is used.
For completeness, we also evaluate our model in Euler angles to compare with the results reported in \cite{gopalakrishnan2019neural}.
To do so, we train our model with exponential maps as rotational representation and achieve competitive results, as can be seen in Table \ref{tab:npss_scales}.
Although our network is still trained to minimize the 3D Euclidean error and not the Euler angle error, we outperform the other methods for most sequences in case of short and medium term forecasting.
Even for long-term forecasting, our model performs consistently well, while vGRU-d, the best-performing model from \cite{gopalakrishnan2019neural}, struggles when forecasting Discussion sequences.

Table \ref{tab:npss_main} compares our model with other state-of-the art models.
As seen in our experiments in Section \ref{sec:shortterm}, we observe that the trajectory dependency (TD) model~\cite{mao2019learning} is highly competitive for short-term prediction.
However, for longer time horizons its performance drastically decreases while our model performs consistently well and achieves state-of-the-art results for long-term motion prediction.
As expected, CNN~\cite{li2018convolutional} also achieves good results over long time horizons, due to its adversarial training.

% ============================================================
% W A R M  U P
% ============================================================
\subsection{Effects of Warm-up Parameter $w$}
\label{sec:warmup}

\begin{figure}[t]
  \begin{center}
  \includegraphics[width=1\linewidth]{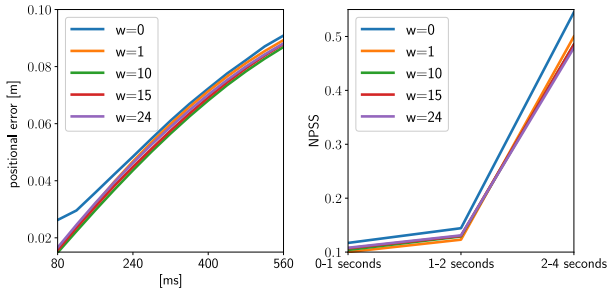}
\end{center}
\caption{
Evaluation of various warm-up periods $w$:
When $w=0$, we observe a strong discontinuity in the first few frames, as can be seen in the left plot.
We also observe that the long-term performance increases slightly with increasing $w$.
}
\label{fig:ablation_w}
\end{figure}

Figure \ref{fig:ablation_w} evaluates various warm-up periods $w$, with the left plot focusing on short-term and the right plot on long-term performance.
No warm-up results in a strong discontinuity on the first few frames but even the setting $w=1$ provides smooth transitions.
In fact, $w\geq 1$ does not have a large influence on short-term motion anticipation performance.
However, long-term prediction seems to improve with larger $w$.
We assume that the warm-up helps the model to reach a desired state faster as it can be seen as a form of partial teacher forcing.

% ============================================================
% C L U S T E R I N G  {K}
% ============================================================
\subsection{Clustering}
\label{sec:clustering_eval}

Our model performs best when the $11$ human-labeled action labels are used.
The classification performance between human-labeled and automatically extracted symbols is almost identical.
The pose label predictor $f_L$ has an accuracy of $88$\% for the $11$ human-labeled classes and an accuracy of $87$\% for unsupervised symbolic representation extraction with $k=11$.
However, the forecasting model $f_F$ achieves a forecast accuracy of $78$\% for $25$ frames anticipation with hand-labeled labels while this number drops to $67$\% when using $k=11$.
On the left plot in Figure \ref{fig:ablation_labels}, we compare several $k$ clustered labels and it becomes evident that the performance decreases with increasing $k$.
This is not so clear for long-term prediction, as can be seen in Table \ref{tab:npss_main}, where $k=32$ performs best for long-term anticipation.

% ================================================
% A B L A T I O N
% ================================================

\subsection{Ablation Studies}

\begin{figure}[t]
  \begin{center}
  \includegraphics[width=1\linewidth]{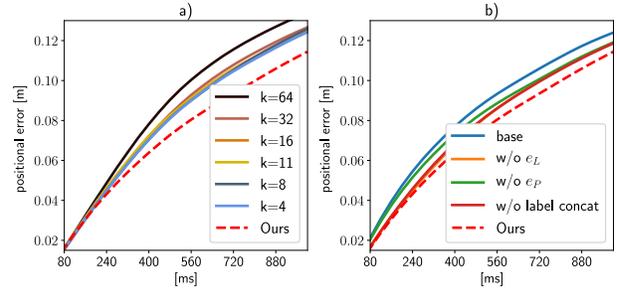}
\end{center}
\caption{
The full model with all submodules is represented as red dotted line.
a) We can see that the unsupervised labels perform worse with increasing $k$.
b) When removing the label concatenation during pose generation in $f_G$ (red line), the label encoder $e_L$ (orange line) or the pose encoder $e_P$ (green line) the forecast performance drops.
The blue line represents the base model $f_G$ without any label or pose encoding and without label concatenation.
}
\label{fig:ablation_labels}
\end{figure}

We evaluate the importance of the various submodules in our human anticipation model.
Removing the concatenation of labels during pose generation or the label encoding both reduce the forecast performance.
Removing the pose encoding especially hinders the early predictions.
Our model performs worst when no labels are used during the generation process and when no label and pose encoding is performed.
The results can be seen in Figure \ref{fig:ablation_labels} b).

\section{Conclusion}

In this work, we presented an approach that forecasts human motion for time horizons between one and four seconds. Instead of forecasting human motion directly from the observed poses, the network forecasts symbolic labels that abstract human motion and uses the observed poses together with the extracted and forecast symbolic labels to forecast the human motion. We evaluated the approach for several protocols on the Human 3.6M dataset where it achieves state-of-the-art results for long-term human motion forecasting. Since our approach is complementary to adversarial training, both ideas can be combined.

{\small
\bibliographystyle{ieee}
\bibliography{egbib}

\begin{thebibliography}{10}\itemsep=-1pt

\bibitem{abu2018will}
Y.~Abu~Farha, A.~Richard, and J.~Gall.
\newblock When will you do what?-anticipating temporal occurrences of
  activities.
\newblock In {\em Conference on Computer Vision and Pattern Recognition}, 2018.

\bibitem{bai2018empirical}
S.~Bai, J.~Z. Kolter, and V.~Koltun.
\newblock An empirical evaluation of generic convolutional and recurrent
  networks for sequence modeling.
\newblock {\em arXiv preprint arXiv:1803.01271}, 2018.

\bibitem{butepage2017deep}
J.~B\"utepage, M.~J. Black, D.~Kragic, and H.~Kjellstrom.
\newblock Deep representation learning for human motion prediction and
  classification.
\newblock In {\em Conference on Computer Vision and Pattern Recognition}, 2017.

\bibitem{farha2019ms}
Y.~A. Farha and J.~Gall.
\newblock Ms-tcn: Multi-stage temporal convolutional network for action
  segmentation.
\newblock In {\em Conference on Computer Vision and Pattern Recognition}, 2019.

\bibitem{fragkiadaki2015recurrent}
K.~Fragkiadaki, S.~Levine, P.~Felsen, and J.~Malik.
\newblock Recurrent network models for human dynamics.
\newblock In {\em International Conference on Computer Vision}, 2015.

\bibitem{gammulle2019forecasting}
H.~Gammulle, S.~Denman, S.~Sridharan, and C.~Fookes.
\newblock Forecasting future action sequences with neural memory networks.
\newblock {\em British Machine Vision Conference}, 2019.

\bibitem{gopalakrishnan2019neural}
A.~Gopalakrishnan, A.~Mali, D.~Kifer, L.~Giles, and A.~G. Ororbia.
\newblock A neural temporal model for human motion prediction.
\newblock In {\em Conference on Computer Vision and Pattern Recognition}, 2019.

\bibitem{gui2018adversarial}
L.-Y. Gui, Y.-X. Wang, X.~Liang, and J.~M. Moura.
\newblock Adversarial geometry-aware human motion prediction.
\newblock In {\em European Conference on Computer Vision}, 2018.

\bibitem{gui2018teaching}
L.-Y. Gui, K.~Zhang, Y.-X. Wang, X.~Liang, J.~M. Moura, and M.~Veloso.
\newblock Teaching robots to predict human motion.
\newblock In {\em International Conference on Intelligent Robots and Systems},
  2018.

\bibitem{holden2016deep}
D.~Holden, J.~Saito, and T.~Komura.
\newblock A deep learning framework for character motion synthesis and editing.
\newblock {\em Transactions on Graphics}, 2016.

\bibitem{h36m_pami}
C.~Ionescu, D.~Papava, V.~Olaru, and C.~Sminchisescu.
\newblock Human3.6m: Large scale datasets and predictive methods for 3d human
  sensing in natural environments.
\newblock {\em Pattern Analysis and Machine Intelligence}, 2014.

\bibitem{jain2016structural}
A.~Jain, A.~R. Zamir, S.~Savarese, and A.~Saxena.
\newblock Structural-rnn: Deep learning on spatio-temporal graphs.
\newblock In {\em Conference on Computer Vision and Pattern Recognition}, 2016.

\bibitem{ke2019time}
Q.~Ke, M.~Fritz, and B.~Schiele.
\newblock Time-conditioned action anticipation in one shot.
\newblock In {\em Conference on Computer Vision and Pattern Recognition}, 2019.

\bibitem{keskar2017improving}
N.~S. Keskar and R.~Socher.
\newblock Improving generalization performance by switching from adam to sgd.
\newblock {\em arXiv preprint arXiv:1712.07628}, 2017.

\bibitem{koppula2015anticipating}
H.~S. Koppula and A.~Saxena.
\newblock Anticipating human activities using object affordances for reactive
  robotic response.
\newblock {\em Transactions on Pattern Analysis and Machine Intelligence},
  2015.

\bibitem{Hodgins:2017:ipend}
T.~Kwon and J.~Hodgins.
\newblock Momentum-mapped inverted pendulum models for controlling dynamic
  human motions.
\newblock {\em Transactions on Graphics}, 2017.

\bibitem{lan2014hierarchical}
T.~Lan, T.-C. Chen, and S.~Savarese.
\newblock A hierarchical representation for future action prediction.
\newblock In {\em European Conference on Computer Vision}, 2014.

\bibitem{li2018convolutional}
C.~Li, Z.~Zhang, W.~Sun~Lee, and G.~Hee~Lee.
\newblock Convolutional sequence to sequence model for human dynamics.
\newblock In {\em Conference on Computer Vision and Pattern Recognition}, 2018.

\bibitem{Hodgins:2017:DOE}
J.~H. Libin~Liu.
\newblock Learning basketball dribbling skills using trajectory optimization
  and deep reinforcement learning.
\newblock {\em Transactions on Graphics}, August 2018.

\bibitem{Hodgins:2017:shedulefrag}
L.~Liu and J.~Hodgins.
\newblock Learning to schedule control fragments for physics-based characters
  using deep q-learning.
\newblock {\em Transactions on Graphics}, 2017.

\bibitem{liu2019towards}
Z.~Liu, S.~Wu, S.~Jin, Q.~Liu, S.~Lu, R.~Zimmermann, and L.~Cheng.
\newblock Towards natural and accurate future motion prediction of humans and
  animals.
\newblock In {\em Conference on Computer Vision and Pattern Recognition}, 2019.

\bibitem{mao2019learning}
W.~Mao, M.~Liu, M.~Salzmann, and H.~Li.
\newblock Learning trajectory dependencies for human motion prediction.
\newblock {\em International Conference on Compuver Vision}, 2019.

\bibitem{martinez2017human}
J.~Martinez, M.~J. Black, and J.~Romero.
\newblock On human motion prediction using recurrent neural networks.
\newblock In {\em Conference on Computer Vision and Pattern Recognition}, 2017.

\bibitem{oord2016wavenet}
A.~v.~d. Oord, S.~Dieleman, H.~Zen, K.~Simonyan, O.~Vinyals, A.~Graves,
  N.~Kalchbrenner, A.~Senior, and K.~Kavukcuoglu.
\newblock Wavenet: A generative model for raw audio.
\newblock {\em arXiv preprint arXiv:1609.03499}, 2016.

\bibitem{paden2016survey}
B.~Paden, M.~{\v{C}}{\'a}p, S.~Z. Yong, D.~Yershov, and E.~Frazzoli.
\newblock A survey of motion planning and control techniques for self-driving
  urban vehicles.
\newblock {\em Transactions on Intelligent Vehicles}, 2016.

\bibitem{pavllo2019modeling}
D.~Pavllo, C.~Feichtenhofer, M.~Auli, and D.~Grangier.
\newblock Modeling human motion with quaternion-based neural networks.
\newblock {\em arXiv preprint arXiv:1901.07677}, 2019.

\bibitem{pavllo:quaternet:2018}
D.~Pavllo, D.~Grangier, and M.~Auli.
\newblock Quaternet: A quaternion-based recurrent model for human motion.
\newblock In {\em British Machine Vision Conference}, 2018.

\bibitem{rubner2000earth}
Y.~Rubner, C.~Tomasi, and L.~J. Guibas.
\newblock The earth mover's distance as a metric for image retrieval.
\newblock {\em International Journal of Computer Vision}, 2000.

\bibitem{ruiz2018human}
A.~H. Ruiz, J.~Gall, and F.~Moreno-Noguer.
\newblock Human motion prediction via spatio-temporal inpainting.
\newblock {\em International Conference on Computer Vision}, 2019.

\bibitem{starke2019nsm}
S.~Starke, H.~Zhang, T.~Komura, and J.~Saito.
\newblock Neural state machine for character-scene interactions.
\newblock {\em Transactions on Graphics}, 2019.

\bibitem{sutskever2014sequence}
I.~Sutskever, O.~Vinyals, and Q.~V. Le.
\newblock Sequence to sequence learning with neural networks.
\newblock In {\em Advances in Neural Information Processing Systems}, 2014.

\bibitem{zhou2018auto}
Y.~Zhou, Z.~Li, S.~Xiao, C.~He, Z.~Huang, and H.~Li.
\newblock Auto-conditioned recurrent networks for extended complex human motion
  synthesis.
\newblock {\em International Conference on Learning Representations}, 2018.

\end{thebibliography}
}

\end{document}